\documentclass[sigconf]{acmart}

\AtBeginDocument{%
  }

\setcopyright{acmcopyright}
\copyrightyear{2018}
\acmYear{2018}
\acmDOI{XXXXXXX.XXXXXXX}

\acmConference[Conference ACM MM '23]{Make sure to enter the correct
  conference title from your rights confirmation email}{Oct. 29--Nov. 03,
  2023}{Ottawa, Canada}
\acmPrice{15.00}
\acmISBN{978-1-4503-XXXX-X/18/06}

\acmSubmissionID{2338}

\def\eg{{\itshape e.g.}}
\def\ie{{\itshape i.e.}}

\def\Ours{{Points-to-3D}}

\begin{document}

%%
%% The "title" command has an optional parameter,
%% allowing the author to define a "short title" to be used in page headers.
%\title{Bridging the Gap between Sparse Points and 3D generation via Diffusion Prior}
%\title{Bridging the Gap between Sparse Points and Shape-Controllable 3D Generation via Diffusion Priors}
\title{Points-to-3D: Bridging the Gap between Sparse Points and Shape-Controllable Text-to-3D Generation}

\author{Chaohui Yu,~~~~~~~~~~Qiang Zhou,~~~~~~~~~~Jingliang Li,~~~~~~~~~~Zhe Zhang,~~~~~~~~~~Zhibin Wang,~~~~~~~~~~Fan Wang\\
DAMO Academy, Alibaba Group\\}
%{\tt \{huakun.ych,jianchong.zq,lijingliang.ljl,qiwu.zz,zhibin.waz,fan.w\}@alibaba-inc.com}}

%%
%% By default, the full list of authors will be used in the page
%% headers. Often, this list is too long, and will overlap
%% other information printed in the page headers. This command allows
%% the author to define a more concise list
%% of authors' names for this purpose.
\renewcommand{\shortauthors}{Trovato et al.}

%之前的方法有2个问题：
%   1.SDS 容易出现多面问题, Janus problem，也就是view之间artifacts
%   2.形状不可控, 最后的结果可能并不是我们想要的形状.
\begin{abstract}
Text-to-3D generation has recently garnered significant attention, fueled by 2D diffusion models trained on billions of image-text pairs. Existing methods primarily rely on score distillation to leverage the 2D diffusion priors to supervise the generation of 3D models, \eg, NeRF. However, score distillation is prone to suffer the view inconsistency problem, and implicit NeRF modeling can also lead to an arbitrary shape, thus leading to less realistic and uncontrollable 3D generation.
In this work, we propose a flexible framework of \Ours{} to bridge the gap between sparse yet freely available 3D points and realistic shape-controllable 3D generation by distilling the knowledge from both 2D and 3D diffusion models. 
%引入3D diffusion model的sparse points先验，和2D latent diffusion model，引入ControlNet的可控生成能力，condition在text 和 depth。
The core idea of \Ours{} is to introduce controllable sparse 3D points to guide the text-to-3D generation. Specifically, we use the sparse point cloud generated from the 3D diffusion model, Point-E, as the geometric prior, conditioned on a single reference image. To better utilize the sparse 3D points, we propose an efficient point cloud guidance loss to adaptively drive the NeRF's geometry to align with the shape of the sparse 3D points.
In addition to controlling the geometry, we propose to optimize the NeRF for a more view-consistent appearance. To be specific, we perform score distillation to the publicly available 2D image diffusion model ControlNet, conditioned on text as well as depth map of the learned compact geometry.
%
% Qualitative and quantitative comparisons demonstrate that \Ours{} improves realism by significantly reducing artifacts among different views and achieves good controllability of 3D shapes for text-to-3D generation.
Qualitative and quantitative comparisons demonstrate that \Ours{} improves view consistency and achieves good shape controllability for text-to-3D generation.
%
%For instance, user studies show \textcolor{red}{xx\%} users to prefer our approach.
%
\Ours{} provides users with a new way to improve and control text-to-3D generation.

\end{abstract}

%%
%% The code below is generated by the tool at http://dl.acm.org/ccs.cfm.
%% Please copy and paste the code instead of the example below.
%%


%%
%% Keywords. The author(s) should pick words that accurately describe
%% the work being presented. Separate the keywords with commas.
%% A "teaser" image appears between the author and affiliation
%% information and the body of the document, and typically spans the
%% page.
\begin{teaserfigure}
  \centering
  \includegraphics[width=0.98\textwidth]{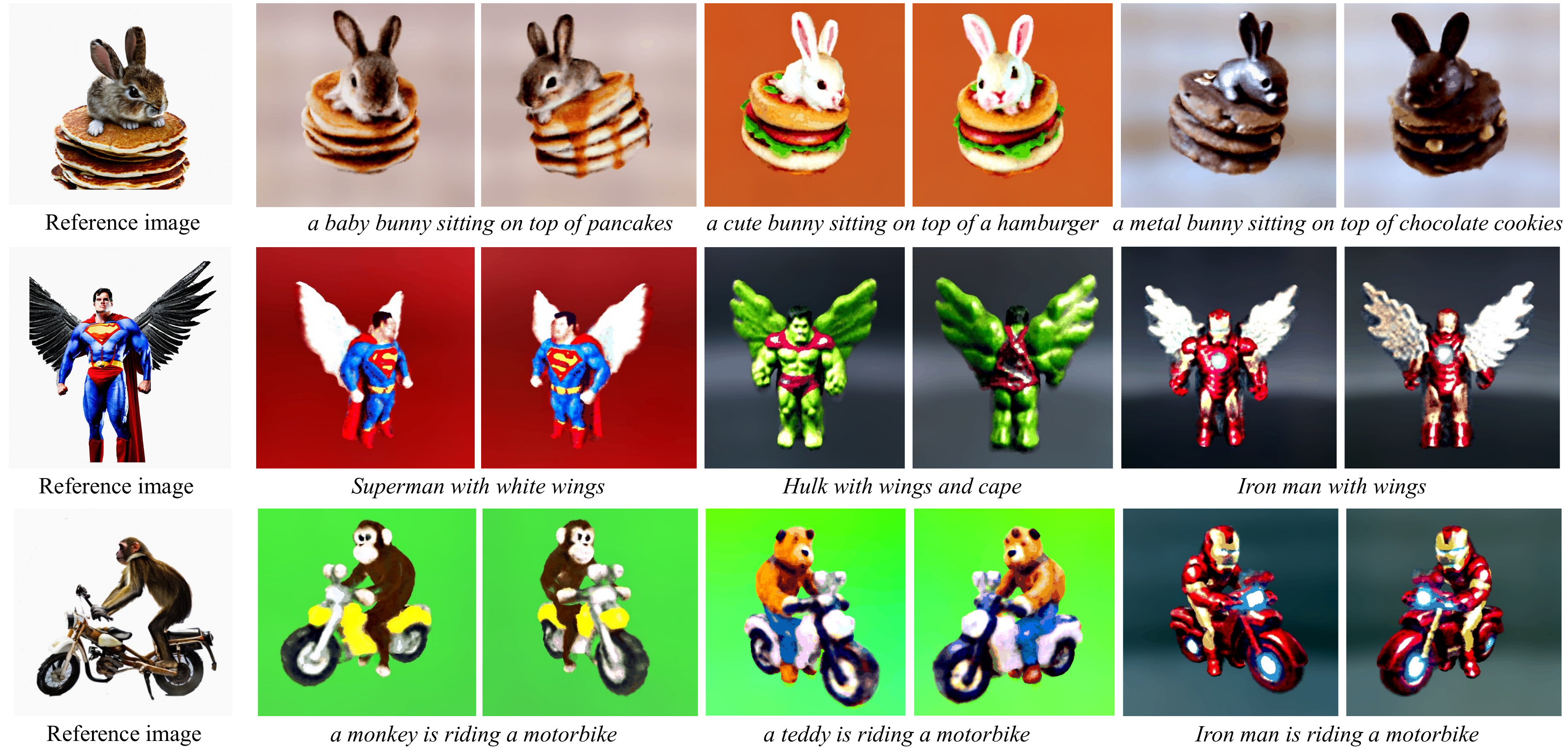}
  \vspace{-.2in}
  \caption{\textit{\Ours{}} can create flexible 3D content with a similar shape to a single reference image. The provided reference image can be a real image or a synthesized image generated by text-to-image diffusion models, \eg, Stable Diffusion.}
  \Description{This is a teaser image. Our method, Points-to-3D can create flexible 3D content with a similar shape to a single reference image.}
  \label{fig:teaser}
\end{teaserfigure}

% NOTE
% \received{20 February 2007}
% \received[revised]{12 March 2009}
% \received[accepted]{5 June 2009}

%%
%% This command processes the author and affiliation and title
%% information and builds the first part of the formatted document.
\maketitle

%################# Introduction #################%
\section{Introduction}
%1. text-to-2d image 生成取得巨大进展，现象级，驱动思考3D怎么生成
Recently, phenomenal advancements have been made in the field of text-to-image generation~\cite{ramesh2021zero,dalle2,Imagen,stablediffusion,controlnet}, mainly due to the significant achievements in large aligned image-text datasets~\cite{laion5b}, vision-language pre-training models~\cite{jia2021scaling,CLIP,BLIP}, and diffusion models~\cite{dhariwal2021diffusion,ddpm,stablediffusion}. Inspired by these text-to-image generation results, many works have explored text-conditional diffusion models in other modalities, \eg, text-to-video~\cite{cogvideo,makeavideo,ho2022video} and text-to-3D~\cite{dreamfields,dreamfusion,magic3d,latentnerf,SJC}. In this work, we focus specifically on the field of text-to-3D generation, which aims to create 3D content and can potentially be applied to many applications, \eg, gaming, virtual or augmented reality, and robotic applications.

%2. text-to-3D最大的问题是没有数据，DreamFusion首次提出SDS 去蒸馏text-to-2d diffusion模型的知识，后面很多方法大多延续着这种做法，比如latent-nerf、SJC，不过这些方法存在两大问题：
%   1.过于隐式训nerf过程中，导致导致生成的模型存在view inconsistency问题，也就是Janus problem，
%   2.同时，由于没有先验指导，导致无法控制比如学出的模型的shape大概是什么样子的。 Latent-nerf第一个提出使用人工设计的mesh 作为先验shape进行指导，但是不generalable，需要对每个场景都要先验设计mesh。
Training text-to-3D generative models can be challenging since it is difficult to attain plentiful text and 3D data pairs compared to 2D images. Most recently, DreamFusion~\cite{dreamfusion} first addresses the challenge by using score distillation from a pre-trained 2D text-to-image diffusion model~\cite{Imagen} to optimize a Neural Radiance Fields (NeRF)~\cite{nerf} to perform text-to-3D synthesis. The following literatures~\cite{latentnerf,SJC} also use the score distillation paradigm. These methods provide and verify the solution for text-to-3D content generation without requiring 3D supervision. Despite their considerable promise, these methods are plagued by a notable issue known as the multi-face problem, or \textit{Janus problem}, which results in inconsistencies across views. 
Besides, another important issue in text-to-3D generation is the lack of control over the shape of the generated 3D objects, \ie, these methods may produce objects with arbitrary shapes that meet the requirements of the input text by setting different seeds. Latent-NeRF~\cite{latentnerf} first introduces sketch-shape guided 3D generation, which uses a predefined mesh as a target to supervise the geometry learning of the NeRF. However, this approach is costly and time-consuming, as it requires the predefinition of a mesh shape for each 3D generation every time.

%3. 本文提出一种新的text-to-3D框架，core idea是 The core idea of \Ours{} is to introduce controllable 3D sparse points to guide the text-to-3D generation. 我们是第一个尝试Bridging the Gap between Sparse Points and Shape-Controllable 3D Generation via 2D and 3D Diffusion Priors. 具体来说:
%   1.提出引入3D 点云diffusion model 针对text或image生成稀疏点云，值得注意的是，这些点云是稀疏的 4096，指导nerf的geometry学习，即density field，并且设计了一个 pointcloud loss
%   2.appearance 通过使用controlnet-depth，与geometry更一致和可控.
This has motivated us to explore the possibility of cultivating prior knowledge in both 2D and 3D diffusion models to guide both the appearance and geometry learning of text-to-3D generation. Inspired by the conditional control paradigm in text-to-image diffusion models, \eg, ControlNet~\cite{controlnet} and T2I-Adapter~\cite{t2i-adapter}, which use extra conditions (\eg, sketch, mask, depth) with text prompts to guide the generation process, achieving more controllability and spatial consistency of the image. We seek a way to incorporate this conditional control mechanism into text-to-3D generation.

In this work, we propose a novel and flexible framework, dubbed \Ours{}, to improve view consistency across views and achieve flexible controllability over 3D shapes for text-to-3D generation. The core idea of \Ours{} is to introduce controllable sparse 3D points to guide the text-to-3D generation in terms of geometry and appearance. To achieve this, inspired by Point-E~\cite{pointE}, we propose to distill the sparse point clouds from pre-trained 3D point cloud diffusion models as the geometry prior. These sparse 3D points are conditioned on a single reference image, which can be provided either by the user or generated by a text-to-image model. However, it is not trivial to leverage the generated sparse point clouds, which only contain 4096 3D points. To overcome this issue, we propose a point cloud guidance loss to encourage the geometry of a randomly initialized NeRF to closely resemble the shape depicted in the reference image. 

In addition to geometry, we propose to optimize the appearance conditioned on text prompt as well as the learned depth map. More concretely, we perform score distillation~\cite{dreamfusion,latentnerf} to the publicly available and more controllable 2D image diffusion models, ControlNet~\cite{controlnet}, in a compact latent space. Our approach, \Ours{}, can bridge the gap between sparse 3D points and realistic shape-controllable 3D generation by distilling the knowledge of 2D and 3D diffusion priors.
As depicted in Figure~\ref{fig:teaser}, given an imaginative reference image, \Ours{} can generate realistic and shape-controllable 3D contents that vary with different text prompts.

In summary, the contributions of this paper are as follows:
\begin{itemize}
    \item We present a novel and flexible text-to-3D generation framework, named \Ours{}, which bridges the gap between sparse 3D points and more realistic and shape-controllable 3D generation by distilling the knowledge from pre-trained 2D and 3D diffusion models. 

    \item To take full advantage of the sparse 3D points, we propose an efficient point cloud guidance loss to optimize the geometry of NeRF, and learn geometry-consistent appearance via score distillation by using ControlNet conditioned on text and learned depth map.

    \item Experimental results show that \Ours{} can significantly alleviate inconsistency across views and achieve good controllability over 3D shapes for text-to-3D generation.
    
\end{itemize}

%################# Related Work #################%
\section{Related Work}
%During the booming of large language models~\cite{llm1}, diffusion models, and scene representation methods, recent years have witnessed remarkable progress in text-to-image generation, NeRF, and text-to-3D generation.

\paragraph{\textbf{Text-to-Image Generation}}
Image generation achieves the first breakthrough results when encountering Generative Adversarial Networks (GANs)~\cite{gan,stylegan}, which train a generator to synthesize images that are indistinguishable from real images. Recently, image generation has achieved another phenomenal progress with the development of diffusion models~\cite{sohl2015deep}. With the improvements in modeling~\cite{ddpm,dhariwal2021diffusion}, denoising diffusion models can generate various high-quality images by iteratively denoising a noised image. In addition to image-driven unconditional generative, diffusion models can generate text-conditioned images from text descriptions~\cite{dalle2,Imagen}. The following works propose to add more conditions to text-to-image generation, including semantic segmentation~\cite{stablediffusion}, reference images~\cite{dreambooth}, sketch~\cite{voynov2022sketch}, depth map~\cite{controlnet,t2i-adapter}, and other conditions~\cite{controlnet,t2i-adapter,composer}, which greatly promote the development and application of text-to-image generation.
Driven by the success of text-to-image diffusion models, many works have explored text-conditional diffusion models in other modalities, \eg, text-based manipulation~\cite{instructpix2pix}, text-to-video~\cite{makeavideo,cogvideo}, and text-to-3D~\cite{dreamfusion,latentnerf,magic3d,SJC}. In this work, we focus on the field of text-to-3D generation.

\paragraph{\textbf{Neural Radiance Fields (NeRF)}}
There is plenty of work on 3D scene representation, including 3D voxel grids~\cite{dvgo}, mesh~\cite{get3d}, point clouds~\cite{achlioptas2018learning,luo2021diffusion,mo2019structurenet,zhou20213d}, and implicit NeRF~\cite{nerf,instantnpg}. In recent years, as a series of inverse rendering methods, NeRF-based methods have emerged as an important technique in 3D scene representation, which are capable of synthesizing novel views and reconstructing geometry surface~\cite{nerf,neus,instantnpg}. Specifically, NeRF~\cite{nerf} represents scenes as density and radiance fields with the neural network (MLP), allowing for photorealistic novel view synthesis. However, the computational cost of densely querying the neural network in 3D space is substantial. To improve the efficiency of NeRF, recent research has explored designing hybrid or explicit structures based on NeRF~\cite{tensorf,instantnpg,dvgo} to achieve fast convergence for radiance field reconstruction, as well as accelerating the rendering speed of NeRF~\cite{fastnerf,hedman2021baking,kilonerf}. Most of these methods require multiple views and corresponding camera parameters for training, which can not be always satisfied, especially in novel text-to-3D content generation.
In this work, we view NeRF as a basic scene representation model and focus on devising a new framework for text-to-3D generation.

\paragraph{\textbf{Single Image 3D Reconstruction}}
Various approaches exist for single image 3D reconstruction, which aims at reconstructing the object present in the image. Different formats can be used to represent the reconstructed object, such as voxels~\cite{3d-r2n2,pix2vox}, polygonal meshes~\cite{pixel2mesh++}, point clouds~\cite{fan2017point}, and more recently, NeRFs~\cite{sinnerf,autorf}. However, these methods are typically trained and evaluated on specific 3D datasets~\cite{shapenet}, making generalization to general 3D reconstruction challenging due to the lack of sufficient 3D training data.
Recently, Point-E~\cite{pointE} explores an efficient method for general 3D content generation in the form of point clouds. It first generates a single synthetic image using a pre-trained text-to-image diffusion model, and then produces a sparse (4096 points) 3D point cloud using a point cloud diffusion model, which is conditioned on the generated image. The generalization ability of Point-E is attributed to its training on several millions of 3D data~\cite{pointE}. In this work, we innovatively leverage Point-E as a point cloud foundation model, to provide sparse geometry guidance for more realistic and shape-controllable text-to-3D generation.

\paragraph{\textbf{Text-to-3D Generation}}
In recent times, the progress in text-to-image generation and 3D scene modeling has sparked a growing interest in text-to-3D content generation. Earlier work like CLIP-forge~\cite{clipforge} consists of an implicit autoencoder conditioned on shape codes and a normalizing flow model to sample shape embeddings from textual input. However, it needs 3D training data in voxel representation, which is difficult to scale in real applications. 
PureCLIPNeRF~\cite{pureclipnerf} uses pre-trained CLIP~\cite{CLIP} for guidance with a voxel grid model for scene representation to perform text-to-3D generation without access to any 3D datasets.
%Text2Mesh~\cite{text2mesh} stylizes a provided 3D mesh by predicting color and local geometric details which conform to a target text prompt through differential rendering and CLIP guidance. Likewise, it requires predefined 3D assets (a specialized 3D mesh) during training.
CLIP-Mesh~\cite{clipmesh} presents a method for zero-shot 3D generation using a textual prompt, it also relies on a pre-trained CLIP model that compares the input text with differentiably rendered images of the generated 3D model.
DreamFields~\cite{dreamfields} first proposes to optimize the 3D representation of NeRF~\cite{nerf}, by employing a pre-trained CLIP as guidance as well, such that all rendering views of NeRF are encouraged to match the text prompt.

More recently, DreamFusion~\cite{dreamfusion} proposes to utilize a powerful pre-trained 2D text-to-image diffusion model~\cite{Imagen} to perform text-to-3D synthesis. They propose a Score Distillation Sampling (SDS) loss to supervise the rendered views of 3D objects modeled by NeRF.
The following Stable-DreamFusion~\cite{stable-dreamfusion}, Latent-NeRF~\cite{latentnerf}, and SJC~\cite{SJC} adapt the score distillation to the publicly available and computationally efficient Stable Diffusion model~\cite{stablediffusion}, which apply the diffusion process in a compact latent space and facilitate the development of text-to-3D generation.
% In addition to NeRF representation, the recent work Point-E~\cite{pointE} explores an alternative method for 3D object generation in the form of point clouds. It first generates a single synthetic view using a text-to-image diffusion model, and then produces a sparse 3D point cloud using a point cloud diffusion model which conditions on the generated image.
We build upon these works and propose a flexible \Ours{} framework for text-to-3D generation by bridging the gap between sparse 3D points and more realistic shape-controllable 3D content generation.

%################# Framework #################%
\begin{figure*}[!t]
    \centering
    \includegraphics[width=0.96\linewidth]{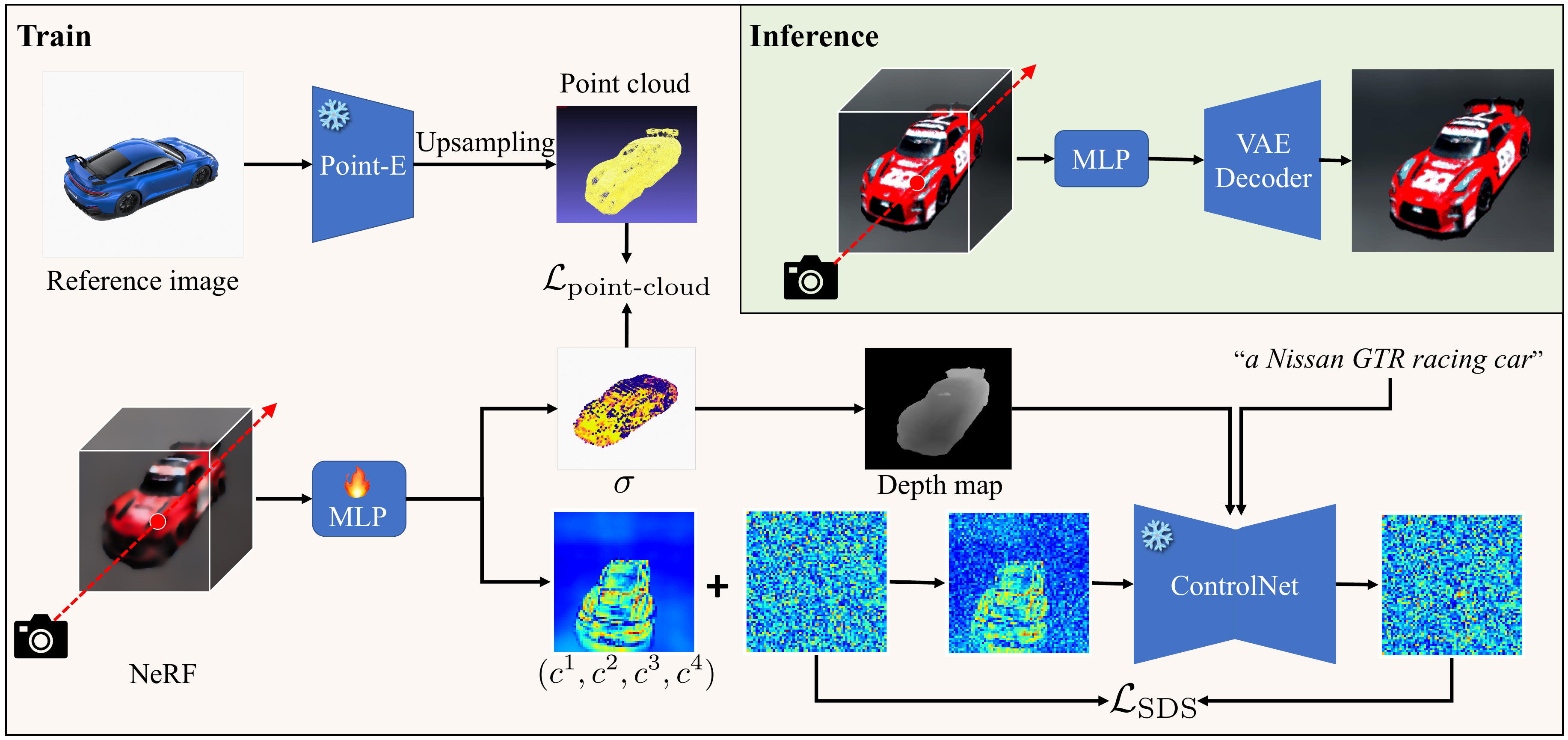}
    \vspace{-.1in}
    \caption{Illustration of the proposed \Ours{} framework for text-to-3D generation. \Ours{} mainly consists of three parts: a scene representation model (a coordinate-based NeRF~\cite{instantnpg}), a text-to-image 2D diffusion model (ControlNet~\cite{controlnet}), and a point cloud 3D diffusion model (Point-E~\cite{pointE}). During training, both 2D and 3D diffusion models are frozen.}
    \Description{This figure shows the framework of our Points-to-3D method, which consists of three parts: a scene representation model, a text-to-image 2D diffusion model, and a point cloud 3D diffusion model. During training, both 2D and 3D diffusion models are frozen.}
    \label{fig:framework}
\end{figure*}

\section{APPROACH}

\subsection{Preliminaries}
In this section, we provide a brief introduction to some of the key concepts that are necessary for understanding our proposed framework in Section~\ref{sec:method}.
%Our method builds off of the latent diffusion model (LDM)~\cite{stablediffusion} and score distillation~\cite{dreamfusion,latentnerf}. In this section, we briefly present a few preliminaries that are necessary for introducing our proposed framework in Section~\ref{sec:method}.

\paragraph{\textbf{Diffusion Model}}
%LDM~\cite{stablediffusion} is a specific form of diffusion models, which 
Diffusion models are first proposed by~\cite{sohl2015deep} and recently promoted by~\cite{song2019generative,ddpm}. A diffusion model usually consists of a forward process $q$ that gradually adds noise to the image $x \in X$, and a reverse process $p$ of gradually removing noise from the noisy data. 
%The forward noising process $q$ is typically a Gaussian distribution and can be formulated as:
The forward process $q$ can be formulated as:
\begin{equation}
q(x_t|x_{t-1}) = \mathcal{N}(x_t; \sqrt{1-\beta_t} x_{t-1}, \beta_t \textbf{I}),
\end{equation}
where timesteps $t \in [0, T]$, $\beta_t$ denotes noise schedule. DDPM~\cite{ddpm} proposes to directly attain a given timestep of the noising procedure:
\begin{equation}
x_t = \sqrt{\bar{\alpha}_t} x_0 + \sqrt{1 - \bar{\alpha}_t} \epsilon,
\end{equation}
where $\bar{\alpha}_t = \prod^{t}_{0}1 - \beta_t$, and $\epsilon \thicksim \mathcal{N}(0, \textbf{I})$.
The denoising process $p_{\theta}(x_{t-1}|x_t)$ starts from random noise and slowly reverses the noising process. DDPM~\cite{ddpm} proposes to parameterize the distribution by modeling the added noise $\epsilon$.
Recently, latent diffusion model (LDM), as a specific form of diffusion model, has achieved great progress in text-to-image generation.
%Since LDM operates on the low-resolution latent space of a pre-trained VAE~\cite{vae}, which is more computationally efficient.
The well-known Stable Diffusion~\cite{stablediffusion} and ControlNet~\cite{controlnet} are both latent diffusion models.

\paragraph{\textbf{Score Distillation Sampling (SDS)}}
%Score distillation sampling (SDS) is first proposed by DreamFusion~\cite{dreamfusion} to achieve text-to-3D generation. To be specific, DreamFusion achieves text-to-3D creation by incorporating two modules: a neural radiance field (NeRF) model (Mip-NeRF 360~\cite{mipnerf}) and a pre-trained 2D text-to-image diffusion model (Imagen~\cite{Imagen}). 
Score distillation sampling (SDS) is first proposed by DreamFusion~\cite{dreamfusion}, which achieves text-to-3D creation by incorporating two modules: a scene representation model~\cite{mipnerf} and a pre-trained text-to-image diffusion model (Imagen~\cite{Imagen}). 
During training, a learnable NeRF model $\theta$ first performs view synthesizes with a differentiable render $g$: $x = g(\theta)$, which can render an image at a given random camera pose. Then, random noise is added to $x$ and the diffusion model $\phi$ is to predict the added noise $\epsilon$ from the noisy image with a learned denoising function $\epsilon_{\phi}(x_t; y, t)$ given the noisy image $x_t$, text embedding $y$, and noise level $t$. This score function provides gradient to update the NeRF parameters $\theta$, which is calculated as:
\begin{equation}
\nabla_{\theta} \mathcal{L}_{\text{SDS}}(\phi, g(\theta)) = \mathbb{E}_{t,\epsilon} \big [ \omega(t)(\epsilon_{\phi}(x_t; y, t) - \epsilon) \frac{\partial x}{\partial \theta} \big ],
\label{eq:sds}
\end{equation}
where $\omega(t)$ is a weighting function that depends on $\alpha_t$.
%Since Imagen~\cite{Imagen} used in DreamFusion is not publicly accessible, we use the publicly available latent diffusion models. 
Inspired by Stable-DreamFusion~\cite{stable-dreamfusion} and Latent-NeRF~\cite{latentnerf}, which use Stable Diffusion~\cite{stablediffusion}, we propose to perform score distillation with a more controllable LDM, ControNet~\cite{controlnet}, to generate more realistic and shape-controllable 3D contents.

\subsection{\Ours{}}
\label{sec:method}
%We inspire our \Ours{} framework by improving the realism among different views and the controllability of 3D shapes for text-to-3D generation. Instead of using a predefined man-made mesh shape as a perfect geometry guidance each time, like the sketch-shape method in Latent-NeRF~\cite{latentnerf}, we propose a novel and flexible method that can leverage the automatically generated sparse 3D points of a pre-trained point cloud diffusion model~\cite{pointE}, conditioning on a user-provided image, as a not-perfect but free geometry guidance. 
In this section, we elaborate on our \Ours{} framework, which is depicted in Figure~\ref{fig:framework}.

\paragraph{\textbf{Architecture}}
First of all, we describe the architecture of our \Ours{} framework. As shown in Figure~\ref{fig:framework}, \Ours{} mainly consists of three models: a scene representation model (a coordinate-based MLP~\cite{instantnpg}), a text-to-image 2D diffusion model (ControlNet~\cite{controlnet}), and a point cloud 3D diffusion model (Point-E~\cite{pointE}). 

$\bullet$ \textbf{Scene Model.} Neural Radiance Field (NeRF)~\cite{nerf} has been an important technique used for scene representation, comprising of a volumetric raytracer and an MLP.
%When rendering a view of a scene, a ray is cast from camera's center of projection through the pixel's location for each pixel in the image plane and out into the world. 
Previous literature~\cite{dreamfusion,latentnerf,SJC} has used NeRF as the scene representation model for text-to-3D generation, mainly because a NeRF model can implicitly impose the spatial consistency between different views owing to the spatial radiance field and rendering paradigm. A NeRF model usually produces a volumetric density $\sigma$ and an RGB color $c$. In this work, we adopt the efficient design of Latent-NeRF~\cite{latentnerf} that produces five outputs, including the volume density $\sigma$ and four pseudo-color channels $\{C = (c^1, c^2, c^3, c^4)\} \in \mathbb{R}^{64 \times 64 \times 4}$ that correspond to the four input latent features for latent diffusion models~\cite{stablediffusion}:
%
% \begin{equation}
% C = \big \{ \sum^N_i w_i c^j_i \big \} \big |^{4}_{j=1},
% \end{equation}
% where
% \begin{equation}
% w_i = \alpha_i \prod_{k<i}(1 - \alpha_k), ~~~~~~~~~~ \alpha_i = 1 - \text{exp}(-\sigma_i || \mu_i - \mu_{i+1} ||).
% \end{equation}
%
\begin{equation}
(c^1, c^2, c^3, c^4, \sigma) = \text{MLP} (x,y,z,d; \theta), 
\end{equation}
where $x,y,z$ denote 3D coordinates, $d$ is the view direction. We use Instant-NGP~\cite{instantnpg} as the scene representation model by default.

$\bullet$ \textbf{Text-to-Image 2D Diffusion Model.} 
Since Imagen~\cite{Imagen} used by DreamFusion~\cite{dreamfusion} is not publicly available, we use Stable Diffusion as the text-to-image diffusion model initially, as previously explored in existing literature~\cite{stable-dreamfusion,latentnerf,SJC}. However, the original Stable Diffusion v1.5 is not controllable to support additional input conditions. In this work, we first propose to use the pre-trained ControlNet~\cite{controlnet} conditioned on depth map as the 2D diffusion model in \Ours{}. As depicted in Figure~\ref{fig:framework}, in addition to the input text prompt, \eg, ``\textit{a Nissan GTR racing car}'', we further introduce the predicted depth map $M \in \mathbb{R}^{H \times W \times 1}$ of our NeRF model as the conditional control. The depth map is computed as follows, for simplicity, we only show the depth value calculation on one pixel:
\begin{equation}
 M_i = \sum^{K}_{k=1} w_k t_k,
 \label{eq:depth}
\end{equation}
and
\begin{equation}
w_k = \alpha_k \prod_{j<k}(1 - \alpha_j), ~~~~\text{and}~~~~ \alpha_k = 1 - \text{exp}(-\sigma_k || t_k - t_{k+1} ||).
\end{equation}
where $K$ is the total number of sampling points along a ray, and $t_k$ denotes the depth hypothesis at point $k$. 
The better and more accurate the predicted depth map $M$, the more geometrically consistent views ControlNet will synthesize.

$\bullet$ \textbf{Point Cloud 3D Diffusion Model.} To control the geometry of NeRF for text-to-3D generation, we propose in this paper, for the first time, the distillation of prior knowledge from the pre-trained large point cloud diffusion model, Point-E~\cite{pointE}. Point-E~\cite{pointE} is an efficient 3D diffusion model for generating sparse 3D point clouds (4096 points) from text prompts or images in about 1 minute. 
As illustrated in Figure~\ref{fig:framework}, we utilize the pre-trained Point-E model to regulate the geometry learning of NeRF. Specifically, the model generates a sparse 3D point cloud consisting of 4096 points, which is conditioned on a reference image and can flexibly represent the object's shape depicted in the image.
However, it is not trivial to guide the NeRF's geometry with only sparse 3D points, we propose a sparse point cloud guidance loss $\mathcal{L}_{\text{point-cloud}}$ to address this issue, which is illustrated in the next section.

It is worth noting that \Ours{} enables users to easily control the shape of the generated content by providing a reference image, which can be a real image or a generated image via text-to-image models~\cite{stablediffusion,controlnet,t2i-adapter}.

\paragraph{\textbf{Sparse 3D Points Guidance}} 
The core idea of our \Ours{} is to introduce controllable sparse 3D points to guide the text-to-3D generation. In this section, we elaborate on how to leverage the sparse 3D points.
%Latent-NeRF uses a predefined and closed mesh, which can naturally divide the $\sigma$ field into two parts: the inner part and the outer part of the mesh.
It is challenging to use a sparse 3D point cloud to guide the geometry learning of NeRF. Previous work on improving NeRF's geometry uses the depth of sparse points to supervise the predicted depth~\cite{densedepthprior,dsnerf}. However, the 3D points are computed using multiple views via COLMAP~\cite{colmap}, and the information about which view each 3D point belongs to has been calculated in advance. 
In our case, only a single RGB image is used to generate the sparse 3D points, when we project all the points to the current view to attain a sparse depth map, there will be aliasing problems between the front and the rear 3D points.

\begin{figure}[!t]
    \centering
    \includegraphics[width=0.96\linewidth]{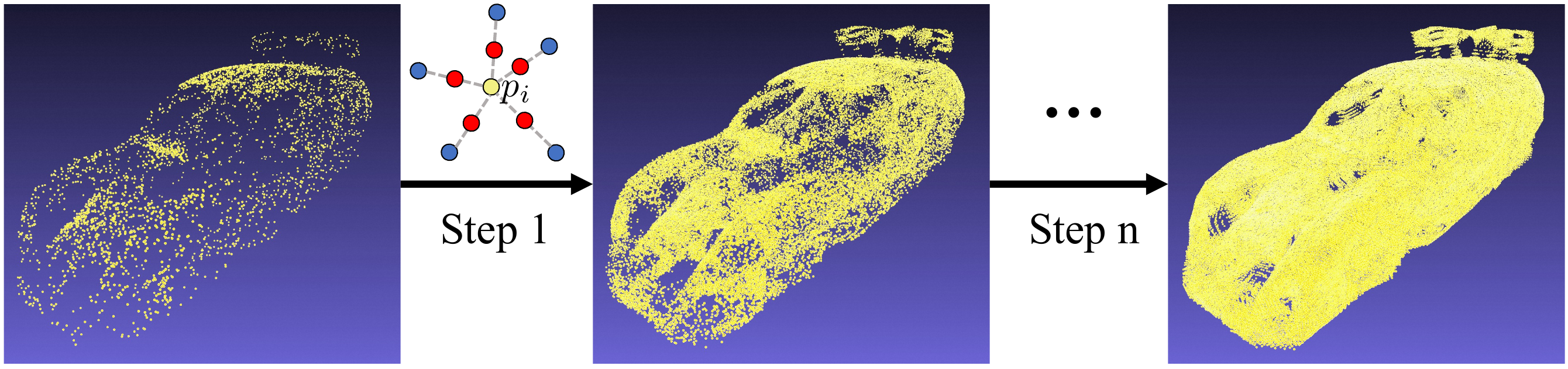}
    \vspace{-.15in}
    \caption{Illustration of the point cloud upsampling process. For each original 3D point (\eg, $p_i$), we add new 3D points (red points) between each of the nearest $q$ neighbor points (blue points) and point $p_i$ for each interpolation step.}
    \Description{This figure shows the point cloud upsampling process. For each original 3D point (\eg, $p_i$), we add new 3D points between each of the nearest $q$ neighbor points and point $p_i$ for each interpolation step.}
    \label{fig:pointup}
    \vspace{-.2in}
\end{figure}

In this work, we present a sparse point cloud guidance loss. Specifically, let $P_s = \{ (x_i,y_i,z_i) \}^{4096}_{i=1}$ be the original sparse 3D points generated by Point-E~\cite{pointE} conditioned on a reference image. Instead of using $P_s$ directly, we experimentally find that making the sparse point cloud to be dense can provide better geometry supervision and produce more realistic 3D contents. We propose to upsample $P_s$ by iteratively performing 3D points interpolation via a simple rule, \ie, for each point $p_i$, we add a new 3D point at the middle position between each of its nearest $q$ neighbor points and $p_i$.
%add a new 3D point at the middle position between each of the nearest $q$ neighbor points and the current point $p_i$. 
The process is depicted in Figure~\ref{fig:pointup}. We set $q = 20, n = 2$ by default. Now we get the dense 3D points $P_d$, which contain about 500k points after eliminating duplicate points. 

Ideally, we want to align the geometry (the volume density $\sigma$) of NeRF with the shape of $P_d$ to ensure that the generated 3D content of \Ours{} closely resembles the reference image.
In addition, we also want to provide NeRF with a level of flexibility and adaptability in its geometry to enable the generation of new details while satisfying different text prompts.
Instead of using the per-view sparse depth map supervision, which has a front-rear aliasing issue as discussed above, and is also not efficient as it only optimizes the current view's depth, we propose an efficient point cloud guidance loss $\mathcal{L}_{\text{point-cloud}}$ to directly optimize the whole geometry ($\sigma$) in 3D space. 
Specifically, we encourage the occupancy ($\alpha$) corresponding to the NeRF points $P_{nerf}$ that near the point cloud $P_d$ to be close to 1, while the occupancy of the NeRF points that far from the point cloud $P_d$ to be close to 0. 
Furthermore, we make the geometry capable of generating new details adaptively by ignoring the supervision of some parts of occupancy. We first compute the closest distance between each point in $P_{nerf}$ and all points in $P_d$: $\mathcal{D} = \mathrm{Dist} (P_{nerf}, P_d)$, $\mathcal{D} \in \mathbb{R}^{S \times 1}$, where $S$ denotes the number of points in $P_{nerf}$. Then, normalize $\mathcal{D}$ via: $\widehat{\mathcal{D}} = \frac{\mathcal{D}}{0.5 \cdot (\max(P_{nerf}) - \min(P_{nerf}))}$. Finally, The calculation of $\mathcal{L}_{\text{point-cloud}}$ can be formulated as:
\begin{equation}
\mathcal{L}_{\text{point-cloud}} = \mathrm{CrossEntropy} (\alpha(P_{nerf}), O(P_{nerf})),
\label{eq:pointcloud}
\end{equation}
and
\begin{equation}
O_i = 
\begin{cases}
1 - \widehat{\mathcal{D}}_i, & \text{if}~ 1 - \widehat{\mathcal{D}}_i > \tau_1;  \\ 
0, & \text{else~if}~ 1 - \widehat{\mathcal{D}}_i < \tau_2;  \\ 
-1, & \text{otherwise};
\end{cases}
\label{eq:gtO}
\end{equation}
where $O(P_{nerf})$ denotes the target occupancy of all NeRF points, $1 - \widehat{\mathcal{D}}$ indicates the degree of proximity to the guided point cloud $P_d$, and $\tau_1, \tau_2$ are two hyperparameters that are experimentally set to 0.95 and 0.9 respectively. We ignore the supervision of points with $\tau_2 < 1 - \widehat{\mathcal{D}} < \tau_1$, allowing the model to adaptively add new details into the geometry to match the text prompts, as well as fix broken holes in the imperfect guided point cloud $P_d$.

\paragraph{\textbf{Training Objectives}}
The training objectives of \Ours{} consist of three parts: the point cloud guidance loss $\mathcal{L}_{\text{point-cloud}}$, the score distillation sampling loss $\mathcal{L}_{\text{SDS}}$, and a sparsity loss $\mathcal{L}_{\text{sparse}}$. 
The sparsity loss is suggested by~\cite{stable-dreamfusion}, which can suppress floaters by regularizing the rendering weights:
\begin{equation}
\mathcal{L}_{\text{sparse}} = -\sum_k (w_k \log w_k + (1 - w_k) \log (1 - w_k)).
\end{equation}
We introduce the depth map condition $M$ calculated by Equation~\ref{eq:depth} and update the score distillation sampling loss in Equation~\ref{eq:sds} as follows:
\begin{equation}
\nabla_{\theta} \mathcal{L}_{\text{SDS}}(\phi, g(\theta)) = \mathbb{E}_{t,\epsilon} \big [ \omega(t)(\epsilon_{\phi}(x_t; y, M, t) - \epsilon) \frac{\partial x}{\partial \theta} \big ].
\label{eq:newsds}
\end{equation}

The overall learning objective is computed as:
\begin{equation}
\mathcal{L} = \lambda_{\text{point}} \mathcal{L}_{\text{point-cloud}} + \lambda_{\text{SDS}} \mathcal{L}_{\text{SDS}} + \lambda_{\text{sparse}} \mathcal{L}_{\text{sparse}}.
\end{equation}

%\paragraph{\textbf{RGB Refinement}}

%################# Experiments #################%
\section{Experiments}
%In the following sections, we conduct a number of ablation studies to verify the effectiveness of our designs and comparisons to show the performance of our \Ours{} in generating more realistic and shape-controllable 3D contents.

\subsection{Baselines}
We consider three text-to-3D generation baselines: DreamFusion~\cite{dreamfusion,stable-dreamfusion}, Latent-NeRF~\cite{latentnerf}, and SJC~\cite{SJC}. 
%DreamFusion~\cite{dreamfusion} first proposes to render 3D objects represented by a NeRF using a pre-trained large text-to-image diffusion model (Imagen~\cite{Imagen}) as guidance.
%Latent-NeRF~\cite{latentnerf} learns a 3D NeRF model on a compact latent feature space ($64 \times 64 \times 4$) instead of in high-resolution (\eg, $512 \times 512$) RGB pixel space. 
%SJC~\cite{SJC}, as a concurrent work to DreamFusion, proposes a similar optimization-based approach to generate 3D assets from 2D diffusion models.
%
Instead of using the close-sourced Imagen~\cite{Imagen} diffusion model, both Latent-NeRF and SJC use the publicly available Stable Diffusion~\cite{stablediffusion}. We mainly compare our \Ours{} with Latent-NeRF and SJC in the experiments. 
We provide more results including comparisons with DreamFields~\cite{dreamfields}, and DreamFusion~\cite{dreamfusion} in our ${\tt supplementary~materials}$.

\subsection{Implementation Details}
%\paragraph{\textbf{NeRF Rendering}}
%We use the code-base provided by Stable-DreamFusion~\cite{stable-dreamfusion}, and 
We use Instant-NGP~\cite{instantnpg} as our scene model. 
%To effectively leverage the latent diffusion models~\cite{stablediffusion,controlnet}, as shown in Figure~\ref{fig:framework}, we follow the design of Latent-NeRF that produces volume density $\sigma$ and four pseudo-color channels $(c^1, c^2, c^3, c^4)$ corresponding to the four latent features that latent diffusion model operates over.
Following the camera sampling method in~\cite{dreamfusion}, during training, a camera position is randomly sampled in spherical coordinates, and we also randomly enlarge the FOV when rendering with NeRF.
%\paragraph{\textbf{Training}}
In addition to the training in latent space shown in Figure~\ref{fig:framework}, we experimentally find that further performing RGB refinement in RGB space, which is introduced in~\cite{latentnerf}, can further improve the text-to-3D generation results.
Our \Ours{} takes less than 50 minutes per text prompt to complete a 3D generation on a single A100 GPU, and most of the time is spent on calculating $\mathcal{L}_{\text{point-cloud}}$. 
We train for 5000 iterations using AdamW optimizer with a learning rate of $1e^{-3}$. The hyperparameters of $\lambda_{\text{point}}, \lambda_{\text{SDS}}, \lambda_{\text{sparse}}$ are set to $5e^{-6}, 1.0, 5e^{-4}$, respectively.

% \paragraph{\textbf{Text-to-Image Diffusion Model.}}
% Inspired by the works that add conditional control in 2D diffusion models, we employ ControlNet~\cite{controlnet} in this work to enable our \Ours{} to better leverage the learned compact geometry guided by sparse 3D points.
% ControlNet~\cite{controlnet} is an expanded work based on Stable Diffusion. It proposes a fast fine-tuning paradigm and reports that large diffusion models like Stable Diffusion can be augmented to enable conditional inputs like depth maps, segmentation maps, keypoints, \etc.
% We use the pre-trained ControlNet conditioned on depth maps in this work.

% \paragraph{\textbf{Point Cloud Diffusion Model.}}
% We use the pre-trained Point-E~\cite{pointE} as the point cloud diffusion model. Point-E~\cite{pointE} is trained on several million processed 3D models from OpenAI. It can generate a 3D object represented as 4096 sparse 3D points using either an input text or a single image as a condition. In our experiments, we use the pre-trained 1B point cloud diffusion model of Point-E.

% \subsection{Metrics}
% FID [24] and KID [5], which are commonly used to evaluate generative models for image synthesis
%借鉴这篇：Generative Novel View Synthesis with 3D-Aware Diffusion Models

\subsection{Ablation Studies}

\begin{figure}[!t]
    \centering
    \includegraphics[width=0.96\linewidth]{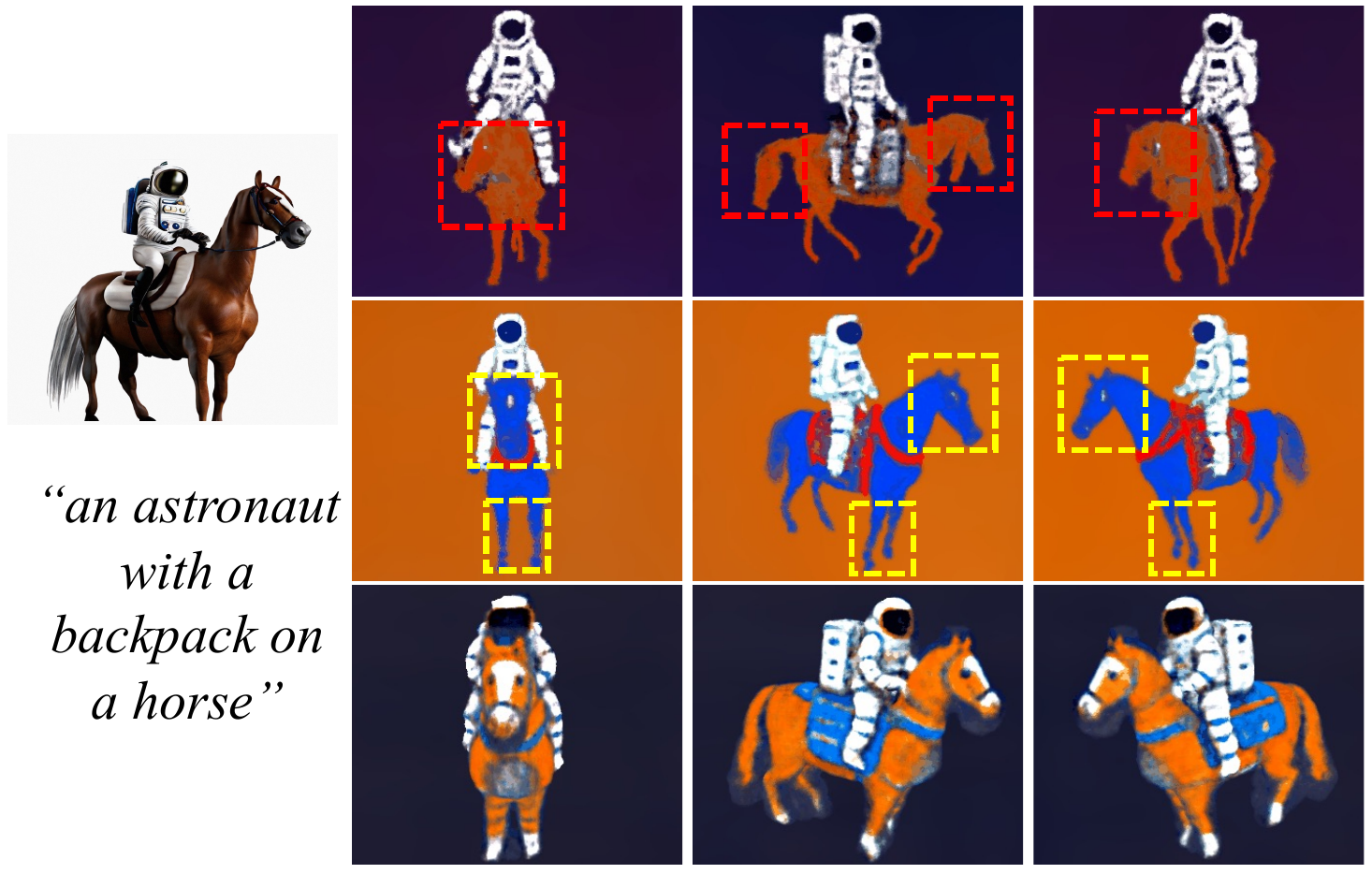}
    \vspace{-.15in}
    \caption{Illustration of the effect of our $\mathcal{L}_{\text{point-cloud}}$. Given a reference image and a text prompt, our \Ours{} with $\mathcal{L}_{\text{point-cloud}}$ (the 3\textit{rd} row) can generate more realistic 3D content than both the per-view depth map loss (the 2\textit{nd} row) and that without any geometry constraints~\cite{latentnerf} (the 1\textit{st} row).}
    \Description{This figure shows the effect of our point cloud guidance loss. Points-to-3D with our points cloud guidance loss (the 3\textit{rd} row) can generate more realistic 3D content than both the per-view depth map loss (the 2\textit{nd} row) and that without any geometry constraints~\cite{latentnerf} (the 1\textit{st} row).}
    \label{fig:pointcloudloss}
    \vspace{-.1in}
\end{figure}

\paragraph{\textbf{Effect of Point Cloud Guidance Loss}}
In this section, we evaluate the proposed point cloud guidance loss $\mathcal{L}_{\text{point-cloud}}$. Concretely, we evaluate \Ours{} by eliminating the point cloud guidance. We also verify the per-view sparse depth map loss as discussed in Section~\ref{sec:method}. The results are shown in Figure~\ref{fig:pointcloudloss}. We first produce a reference image with the text prompt: ``an astronaut with a backpack on a horse'' using Stable Diffusion. Then we use $\mathcal{L}_{\text{point-cloud}}$ (the 3\textit{rd} row), a designed per-view depth map loss (the 2\textit{nd} row), and without any geometry constraints (the 1\textit{st} row), to train three models with the same text prompt, respectively. We can find that without any geometry constraints, 
%which can also be seen as the baseline~\cite{latentnerf}, 
the generated content suffers an obvious view inconsistency problem (red dashed boxes).
The result of using our designed per-view depth map loss as geometry supervision further improves the multi-face issue. However, the rendered images are less realistic and even broken (yellow dashed boxes) due to the sparsity of point clouds and the inefficiency of the per-view supervision. 
It is worth noting that the result of using $\mathcal{L}_{\text{point-cloud}}$ shows more details in both ``astronaut'' and ``horse''. 
That is, \Ours{} with $\mathcal{L}_{\text{point-cloud}}$ for geometry optimization can generate more realistic 3D content.

\begin{figure}[!t]
    \centering
    \includegraphics[width=0.96\linewidth]{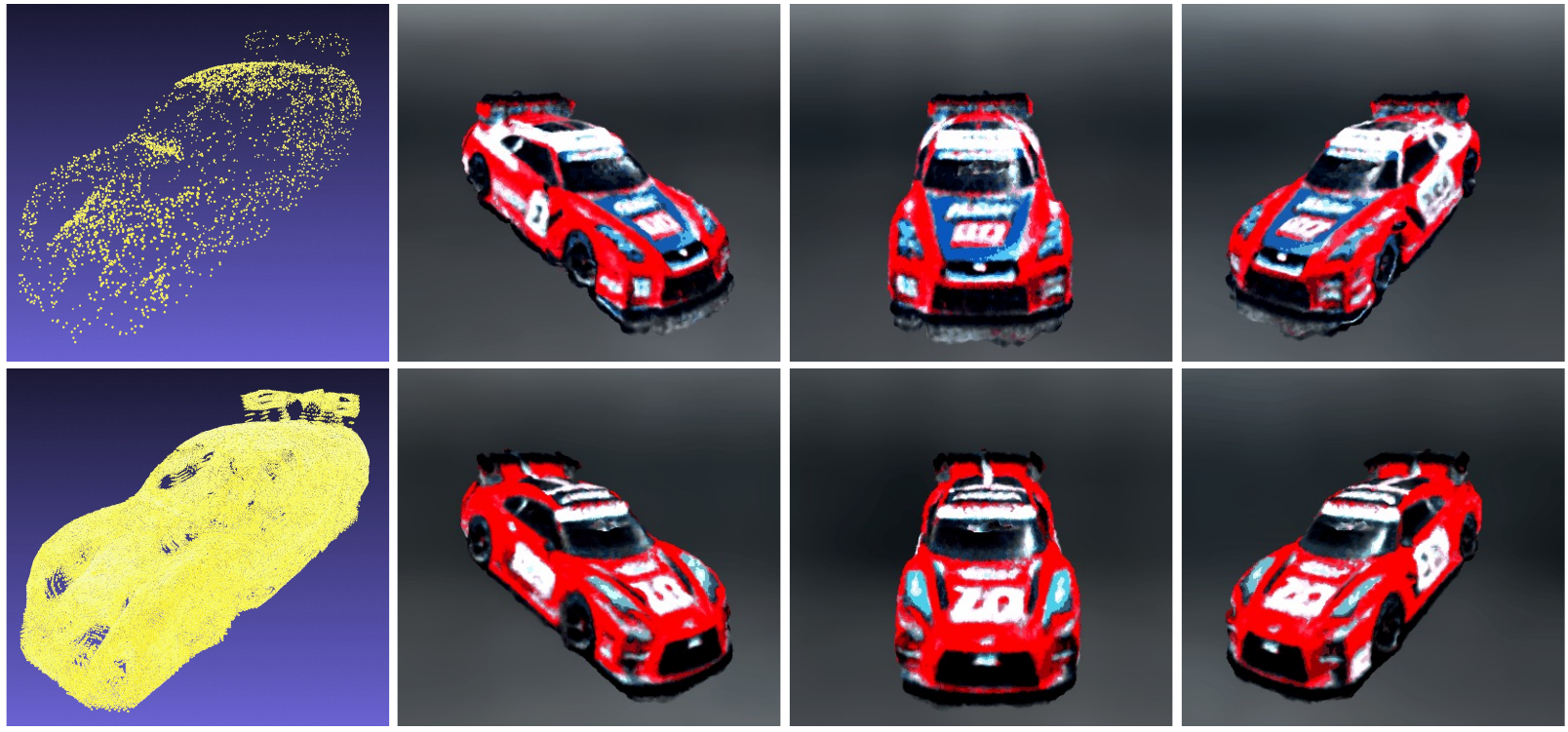}
    \vspace{-.15in}
    \caption{Comparison of rendered views of models trained with $P_s$ and $P_d$ as geometry guidance, respectively. The text prompt is ``a Nissan GTR racing car''.}
    \Description{This figure shows a comparison of rendered views of models trained with $P_s$ and $P_d$ as geometry guidance, respectively.}
    \label{fig:pointupsample}
    \vspace{-.1in}
\end{figure}

\paragraph{\textbf{Effect of 3D Points Upsampling}}
In this section, we analyze the effect of upsampling the generated sparse 3D point cloud. As shown in Figure~\ref{fig:pointupsample}, we compare the rendered views of \Ours{} trained with sparse (4096) 3D points $P_s$ and upsampled denser ($\sim$500k) 3D points $P_d$ as the geometry guidance, respectively. The 1\textit{st} column represents the original sparse points $P_s$ produced by Point-E~\cite{pointE} given the reference image shown in Figure~\ref{fig:framework}, and the upsampled points $P_d$ via our designed rule. The 2\textit{nd}~$\sim$~4\textit{th} columns are three corresponding rendered views. 
We can see that the results guided by $P_d$ are more realistic compared to those guided by $P_s$. This is due to that a denser point cloud can offer more supervision to encourage the NeRF to learn a more concise geometry.
Moreover, better geometry (depth map) can also guide ControlNet~\cite{controlnet} to generate more geometry-consistent and realistic images that match the input text prompt.

\begin{figure}[!t]
    \centering
    \includegraphics[width=0.96\linewidth]{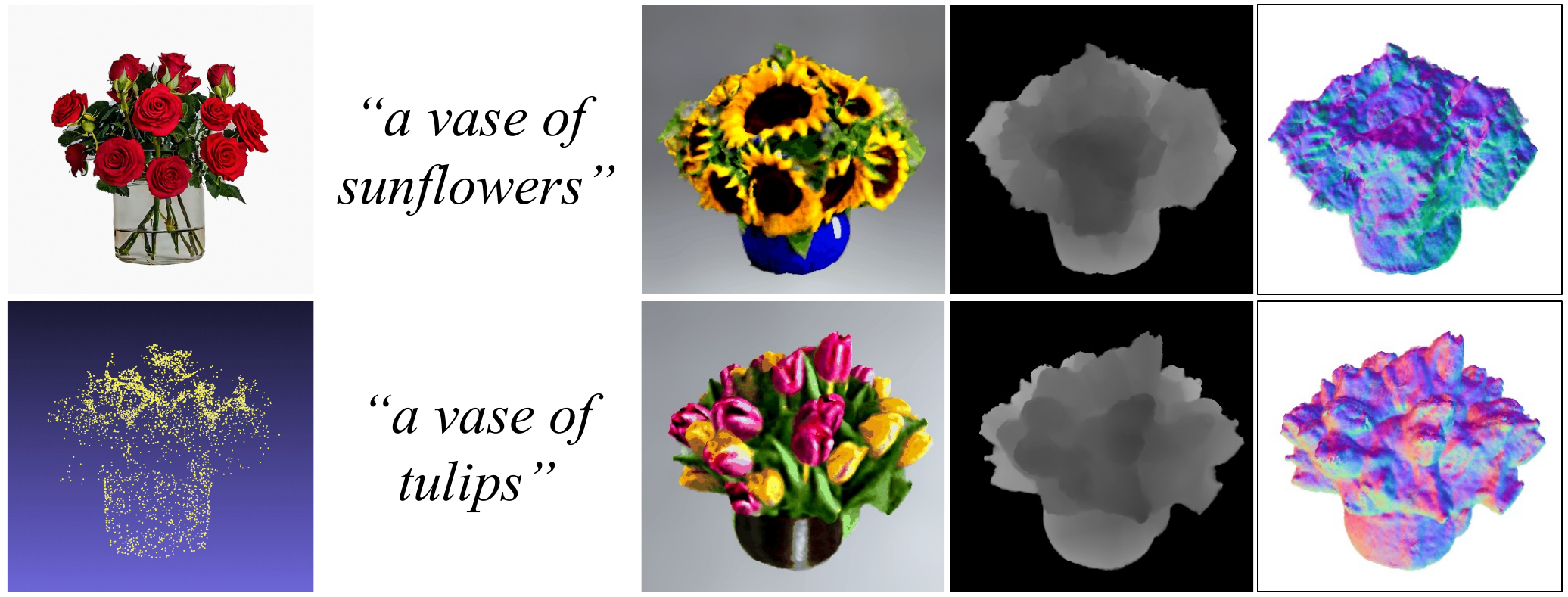}
    \vspace{-.15in}
    \caption{Visualization of two 3D models trained with the same reference image (generated by Stable Diffusion~\cite{stablediffusion}) and the corresponding sparse 3D points but different texts.}
    \Description{This figure shows the visualization of two 3D models trained with the same reference image and the corresponding sparse 3D points but different texts. The last three columns represent the rendered images, the rendered depth maps, and the rendered normals at the same camera pose, respectively.}
    \label{fig:adaptivedesign1}
    \vspace{-.15in}
\end{figure}

\begin{figure}[!t]
    \centering
    \includegraphics[width=0.96\linewidth]{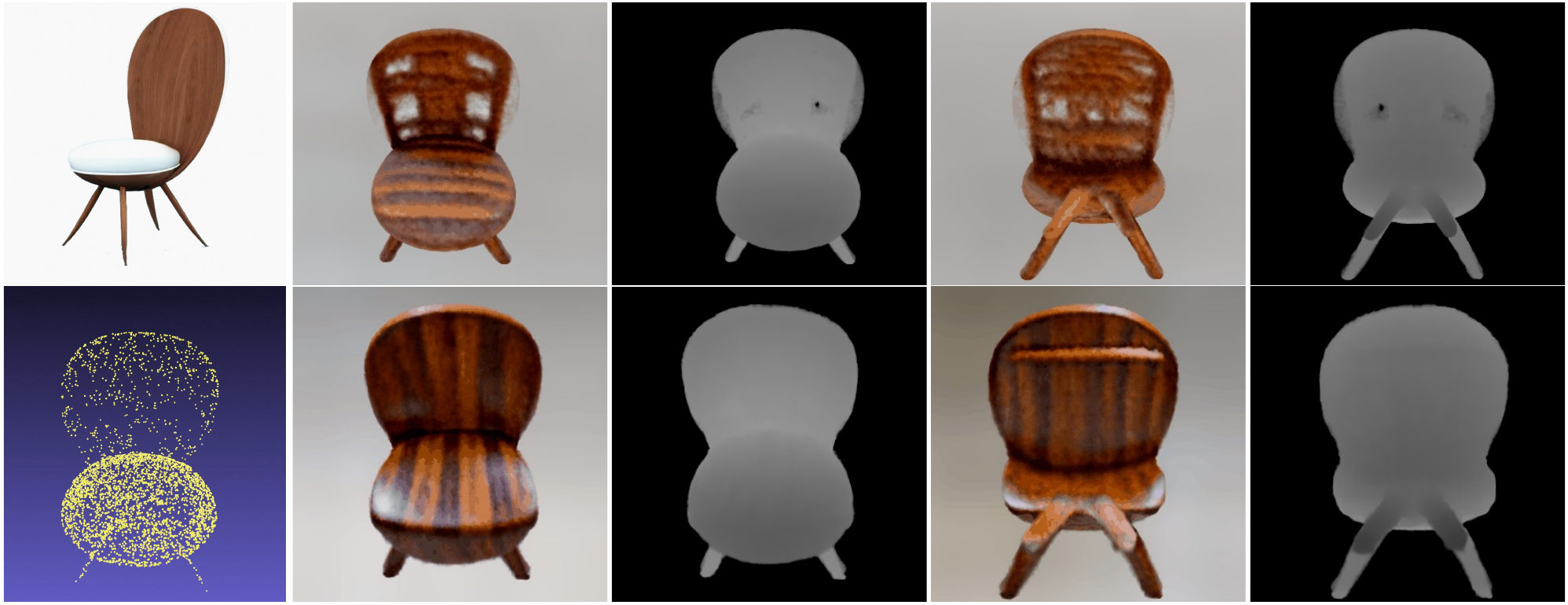}
    \vspace{-.15in}
    \caption{Comparison of two 3D models trained with the same reference image and sparse 3D points shown in the 1\textit{st} column. The 1\textit{st} and the 2\textit{nd} rows denote training without and with adaptive design in $\mathcal{L}_{\text{point-cloud}}$, respectively. The text prompt is ``a wooden chair''.}
    \Description{This figure shows two 3D models trained with the same reference image and sparse 3D points. The 1\textit{st} and the 2\textit{nd} rows denote training without and with adaptive design in $\mathcal{L}_{\text{point-cloud}}$, respectively.}
    \label{fig:adaptivedesign2}
    \vspace{-.1in}
\end{figure}

\paragraph{\textbf{Effect of Adaptive Design in $\mathcal{L}_{\text{point-cloud}}$}}
%1. 自适应填充细节
%2. 填补空洞
In this section, we illustrate the effect of the adaptive design in $\mathcal{L}_{\text{point-cloud}}$. That is, in Equation~\ref{eq:pointcloud} and Equation~\ref{eq:gtO}, we propose to ignore the supervision of those NeRF points with $\tau_2 < 1 - \widehat{\mathcal{D}} < \tau_1$ to let \Ours{} to adaptively adjust the geometry to match the text prompt. This adaptive design serves two main purposes: a). it offers the capacity to create new details without changing the main shape of the 3D content. b). it can fill broken holes in the imperfect point clouds $P_d$.

\begin{figure*}[!t]
    \centering
    \includegraphics[width=1.0\linewidth]{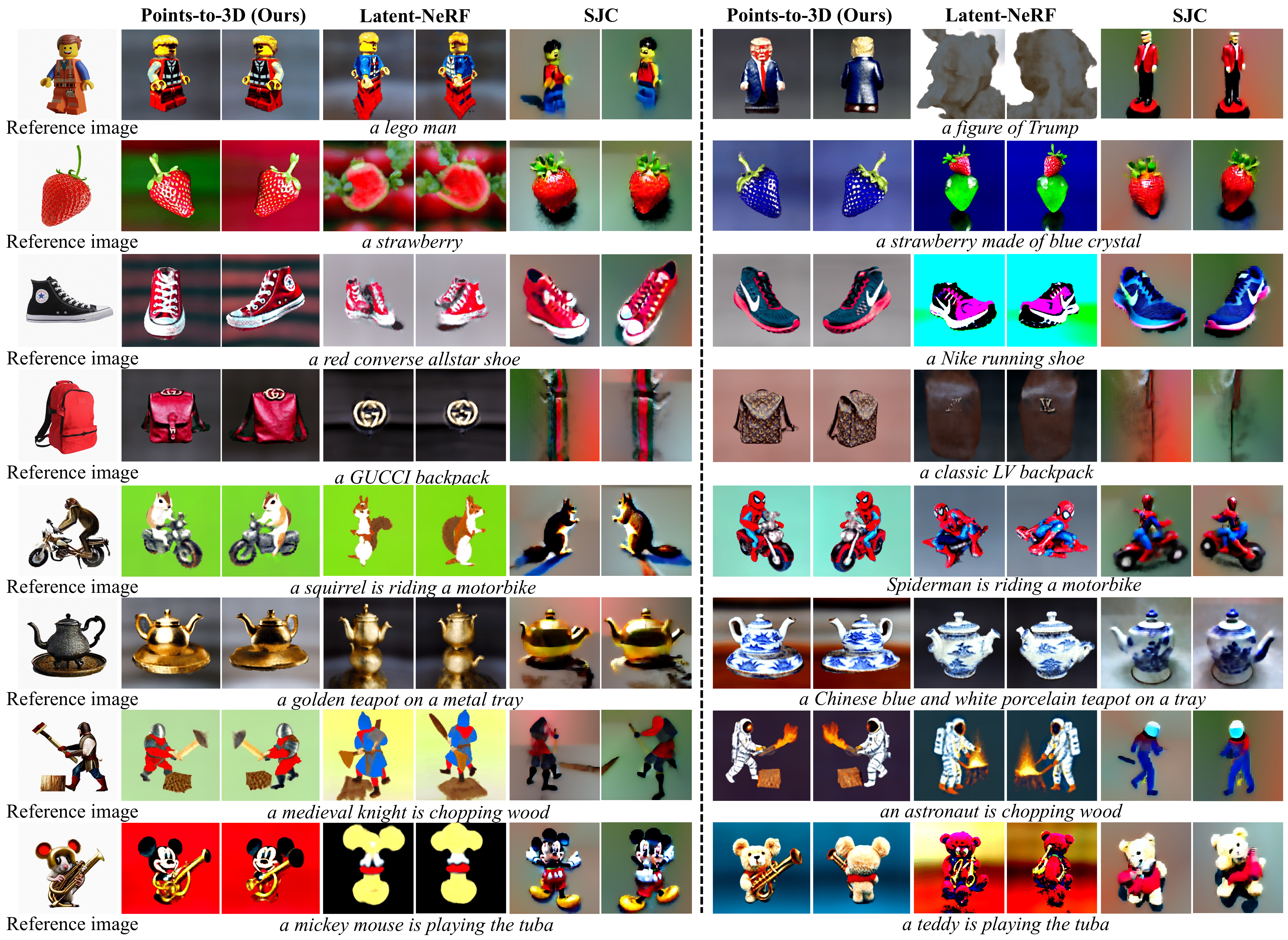}
    \vspace{-.3in}
    \caption{Qualitative comparison with Latnet-NeRF~\cite{latentnerf} and SJC~\cite{SJC} on single-object generation (the 1\textit{st} $\sim$ 4\textit{th} rows) and scene generation (the 5\textit{th} $\sim$ 8\textit{th} rows). The 1\textit{st} column denotes reference images used for \Ours{}, where the top four are real images and the bottom four are synthetic images generated using Stable Diffusion~\cite{stablediffusion}. (Best viewed by zooming in.)}
    \Description{This figure shows a comparison with Latnet-NeRF and SJC on single-object generation (the 1\textit{st} $\sim$ 4\textit{th} rows) and scene generation (the 5\textit{th} $\sim$ 8\textit{th} rows).}
    \label{fig:result-real-synthe}
    \vspace{-.1in}
\end{figure*}

As shown in Figure~\ref{fig:adaptivedesign1}, we visualize two generated 3D contents using \Ours{} with the same reference image and sparse point cloud but different text prompts. The last three columns represent the rendered images, the rendered depth maps, and the rendered normals at the same camera pose, respectively. We can clearly observe that \Ours{} can generate more specific new details to match different input text prompts based on the same point cloud guidance.
In Figure~\ref{fig:adaptivedesign2}, we analyze the effect of adaptive design in filling holes in the imperfect point cloud. Given a reference image, Point-E~\cite{pointE} may produce non-uniform point clouds, \eg, broken holes in the chair back in this instance. If we enforce all the NeRF points closed to the point cloud to be positive class and otherwise negative class, it is difficult to set an appropriate distance threshold for all 3D contents and will cause broken holes. For instance, we compare the results of rendered images and corresponding depth maps trained without and with adaptive design in the 1\textit{st} and 2\textit{nd} row, respectively. \Ours{} can naturally repair the broken holes in both geometry and appearance.
We also analyze the effect of the depth map condition in our ${\tt supplementary~materials}$.

\subsection{Shape-Controllable Text-to-3D Generation}
As special concepts and shapes are usually difficult to describe by text prompts but easy with images, it is desperately needed to have a mechanism to guide the text-to-3D content generation with images. In this section, we evaluate \Ours{} in generating view-consistent and shape-controllable 3D contents with a single reference image for geometry guidance. Considering that DreamFusion~\cite{dreamfusion} and Magic3D~\cite{magic3d} use their proprietary text-to-image diffusion models~\cite{Imagen,ediffi} and neither releases the code, we mainly compare with Latent-NeRF~\cite{latentnerf} and SJC~\cite{SJC}. As shown in Figure~\ref{fig:result-real-synthe}, we mainly compare two aspects: single-object generation and scene (consists of multiple objects) generation. 

%TODO
For the single-object generation (the 1\textit{st} $\sim$ 4\textit{th} rows), 
Latent-NeRF~\cite{latentnerf} is easy to suffer the view inconsistency problem, and sometimes fails to generate reasonable content.
SJC~\cite{SJC} looks a little better than Latent-NeRF in terms of view consistency of the generated objects, however, it also sometimes fails to generate content that matches the text description (\eg, the 2\textit{nd} and the 4\textit{th} rows).
Our \Ours{} can automatically generate view-consistent and more realistic single objects. It is worth noting that \Ours{} can generate more lifelike details, \eg, the logos of Converse, Nike, GUCCI, and LV.

For more challenging scene generation (the 5\textit{th} $\sim$ 8\textit{th} rows), the inherent view inconsistency problem of Latent-NeRF~\cite{latentnerf} becomes more serious, \eg, multiple teapot spouts in the 6\textit{th} row and multiple hands or legs in the 7\textit{th} row. Besides, both Latent-NeRF and SJC can easily lose some concepts of the input text prompts, \eg, ``motorbike'' in the 5\textit{th} row, ``tray'' in the 6\textit{th} row, and ``tuba'' in the last row. In contrast, our \Ours{} can create view-consistent 3D content and preserve the concepts contained in the text prompts.

Furthermore, \Ours{} enables users to arbitrarily create or modify 3D content that has a similar shape to the reference image. We provide more comparisons in our ${\tt supplementary~materials}$.

\begin{figure}[!t]
    \centering
    \includegraphics[width=1.0\linewidth]{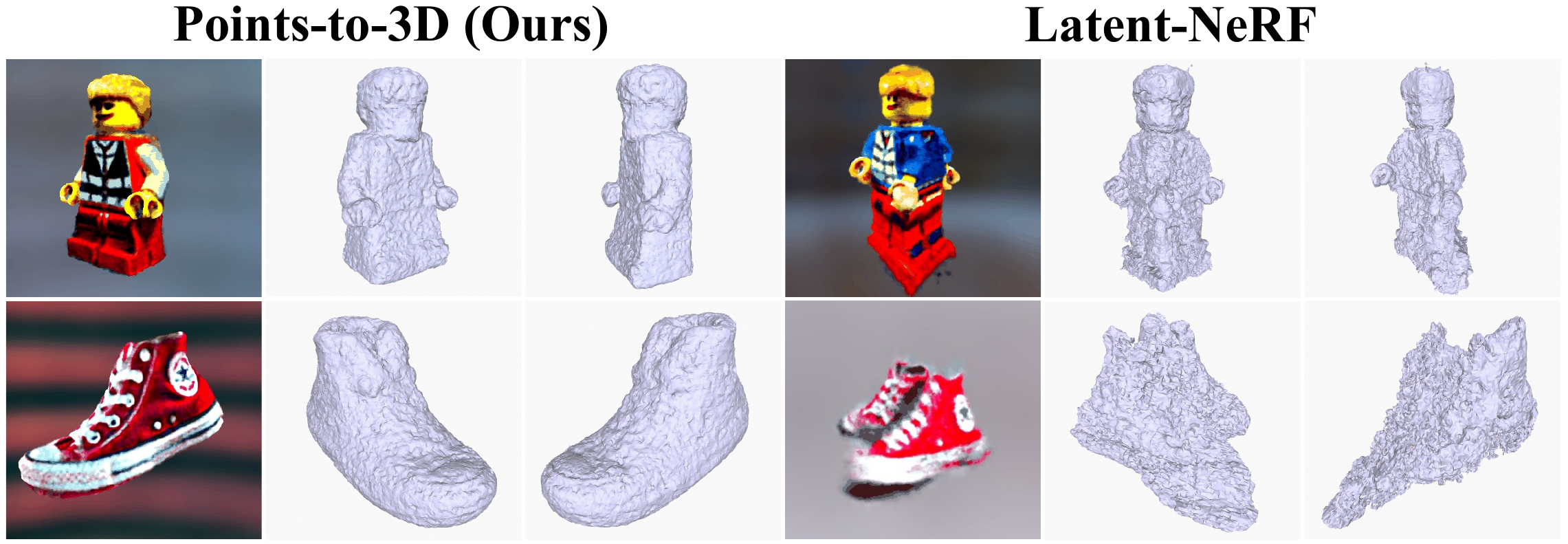}
    \vspace{-.3in}
    \caption{Mesh comparison through Marching Cubes~\cite{marchingcubes}.}
    \Description{This figure shows the mesh comparison through Marching Cubes. Points-to-3D can generate more compact and delicate geometry than Latent-NeRF.}
    \label{fig:result-meshcompare}
    \vspace{-.1in}
\end{figure}

\subsection{Geometry Comparison}
%对比生成的mesh
%In this section, 
We compare the learned geometry of \Ours{} and Latent-NeRF~\cite{latentnerf}, both of which use Instant-NGP~\cite{instantnpg} as the scene model. As depicted in Figure~\ref{fig:result-meshcompare}, we show two generation results produced using two text prompts: ``a lego man'' and ``a red converse allstar shoe''. Each contains three views: a rendered RGB image and two views of mesh. The meshes are extracted by Marching Cubes~\cite{marchingcubes} from density field of the learned Instant-NGP.
We can clearly observe that compared to the flawed meshes of Latent-NeRF, \Ours{} can generate more delicate meshes.
That is, in addition to synthesis view-consistent novel views, \Ours{} can learn controllable and more compact geometry for text-to-3D generation.

\begin{figure}[!t]
    \centering
    \includegraphics[width=1.0\linewidth]{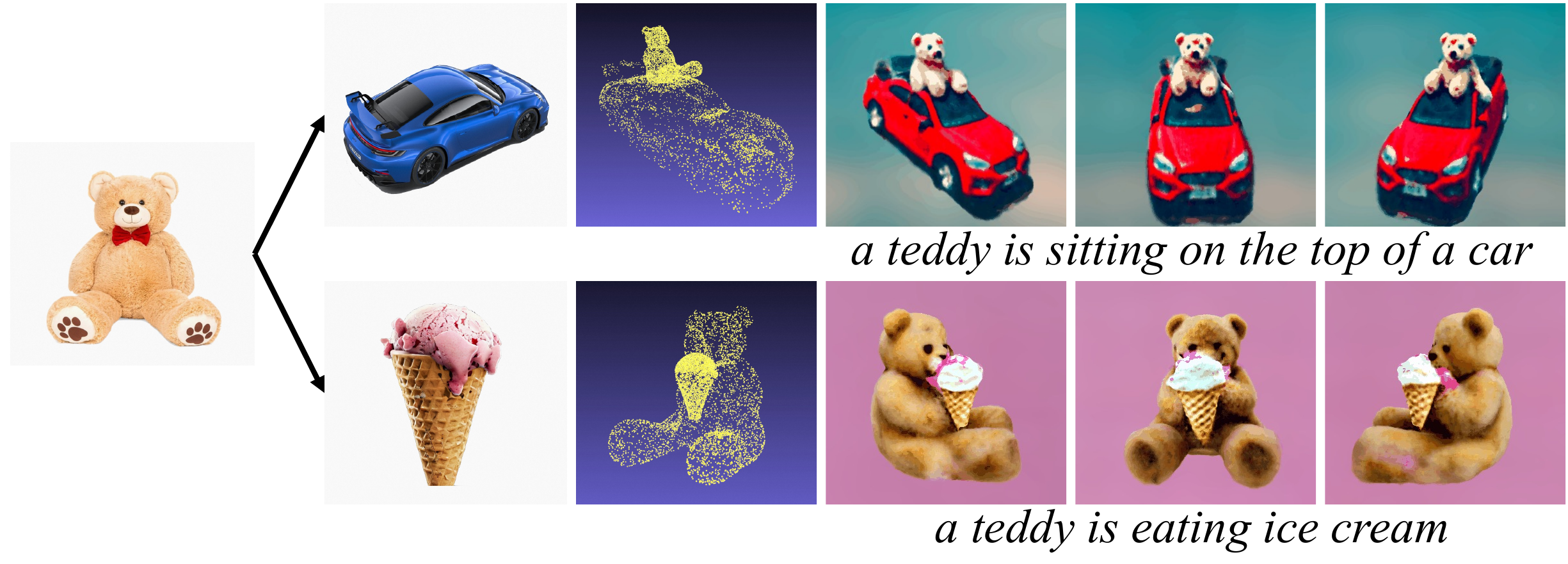}
    \vspace{-.3in}
    \caption{Compositional generation of \Ours{}.}
    \Description{This figure shows that our Points-to-3D framework can flexibly perform compositional generation by using multiple reference images.}
    \label{fig:result-composite}
    \vspace{-.1in}
\end{figure}

\subsection{Compositional Generation}
%两个或多个稀疏点云合起来去作为一个新的指导
We analyze the effectiveness of \Ours{} in generating compositional 3D content. As shown in Figure~\ref{fig:result-composite}, by taking the manually composited sparse 3D points of multiple reference images as geometry guidance, \Ours{} can perform view-consistent and shape-controllable text-to-3D generation. The results indicate that \Ours{} enables users to freely composite objects using multiple reference images and generate more imaginative 3D content.

% \begin{table}[!t]
% \caption{Quantitative comparison using CLIP R-precision of Latent-NeRF~\cite{latentnerf}, SJC~\cite{SJC}, and our \Ours{}.}
% \vspace{-.15in}
% \label{tbl:r_precision}
% \centering
% \resizebox{0.94\linewidth}{!}{
% \begin{tabular}{c|ccc}
% \toprule
% Method                            & ViT-B/16 $\uparrow$ & ViT-B/32 $\uparrow$   & ViT-L/14 $\uparrow$ \\
% \midrule
%     Latent-NeRF~\cite{latentnerf} &        53.00\%           &   59.00\%            &  66.00\%     \\
%     SJC~\cite{SJC}                &        61.00\%           &   57.00\%            &  71.00\%     \\
%     \Ours{} (Ours)                &        \textbf{81.00\%}  &   \textbf{81.00\%}   &  \textbf{90.00\%}     \\
% \bottomrule
% \end{tabular}
% }
% %\vspace{-.1in}
% \end{table}

% \begin{figure}[!t]
%     \centering
%     \includegraphics[width=0.9\linewidth]{figures/result_userstudy.pdf}
%     \vspace{-.15in}
%     \caption{Quantitative comparison via user studies with 22 participants to measure preference in terms of view consistency and prompt relevance.}
%     %\Description{xxx.}
%     \label{fig:result-user_study}
%     \vspace{-.1in}
% \end{figure}

\subsection{Quantitative Comparisons}
%CLIP R-Precision: DreamBooth3D, DreamFusion
%User study: Magic3D, DreamBooth3D
% realistic # shape-controllable # Training speed, 可以搞一个条形图

\paragraph{\textbf{CLIP R-precision}}
In this section, we calculate the CLIP R-precision metrics for Latent-NeRF~\cite{latentnerf}, SJC~\cite{SJC}, and our \Ours{}. We compute CLIP R-precision following~\cite{dreamfields} on 50 text and 3D model pairs (shown in our ${\tt supplementary~materials}$) based on three CLIP image encoders (ViT-B/16, ViT-B/32, and ViT-L/14). For each 3D generation, we randomly select two rendered views for calculation. The results are reported in Table~\ref{tbl:r_precision}, the higher scores for our \Ours{} results indicate that renderings from our 3D model outputs more accurately resemble the text prompts.

\begin{figure}[!t]
    \centering
    \includegraphics[width=0.9\linewidth]{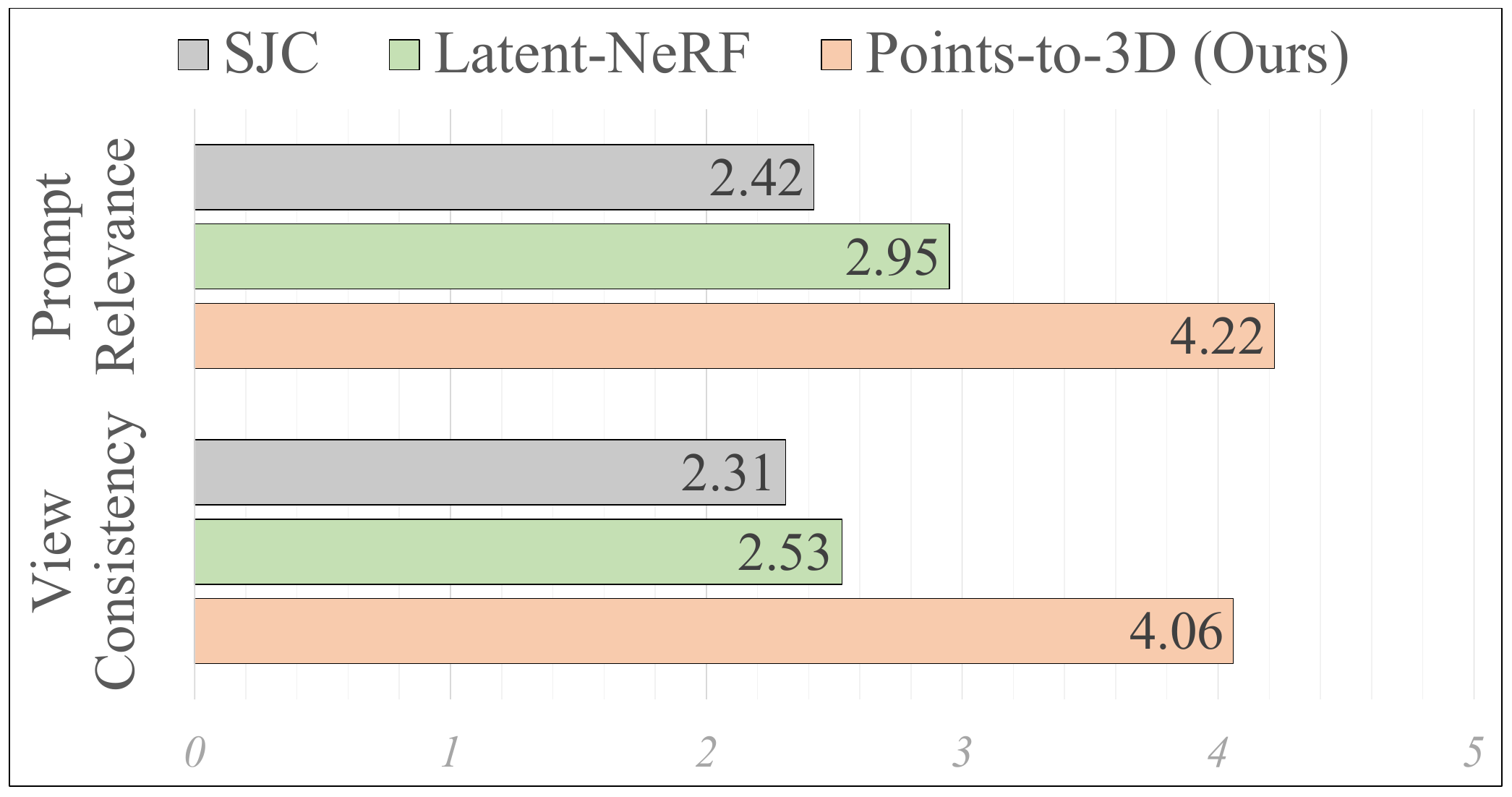}
    \vspace{-.15in}
    \caption{Quantitative comparison via user studies with 22 participants to measure preference in terms of view consistency and prompt relevance.}
    \Description{This figure shows quantitative comparison via user studies with 22 participants to measure preference in terms of view consistency and prompt relevance.}
    \label{fig:result-user_study}
    \vspace{-.1in}
\end{figure}

\paragraph{\textbf{User Studies}}
The CLIP R-precision metric focuses on the matching degree of rendered views and text prompts, but it is difficult to reflect view consistency and image realism.
We conduct user studies with 22 participants to evaluate different methods based on user preferences. We ask the participants to give a preference score (range from 1 $\sim$ 5) in terms of view consistency and prompt relevance for each anonymized method's generation.
As shown in Figure~\ref{fig:result-user_study}, we report the average scores on a randomly composed evaluation set that consists of 36 generation results of each method. We find that \Ours{} is significantly preferred over both Latent-NeRF and SJC in terms of view consistency and prompt relevance.
We provide more detailed information about the user study, please refer to our ${\tt supplementary~materials}$.

\begin{table}[!t]
\caption{Quantitative comparison using CLIP R-precision of Latent-NeRF~\cite{latentnerf}, SJC~\cite{SJC}, and our \Ours{}.}
\vspace{-.15in}
\label{tbl:r_precision}
\centering
\resizebox{0.92\linewidth}{!}{
\begin{tabular}{c|ccc}
\toprule
Method                            & ViT-B/16 $\uparrow$ & ViT-B/32 $\uparrow$   & ViT-L/14 $\uparrow$ \\
\midrule
    Latent-NeRF~\cite{latentnerf} &        53.00\%           &   59.00\%            &  66.00\%     \\
    SJC~\cite{SJC}                &        61.00\%           &   57.00\%            &  71.00\%     \\
    \Ours{} (Ours)                &        \textbf{81.00\%}  &   \textbf{81.00\%}   &  \textbf{90.00\%}     \\
\bottomrule
\end{tabular}
}
\vspace{-.1in}
\end{table}

%################# Discussion #################%
\section{Limitations}

%While \Ours{} allows for flexible text-to-3D generation with a similar shape to a single reference image and improves over prior works in terms of realism, view consistency, and shape-controllable, we observe several limitations.
While \Ours{} allows for flexible text-to-3D generation and improves over prior works in terms of realism, view consistency, and shape controllability, we observe several limitations. 
First, as \Ours{} is built upon pre-trained 2D image diffusion model~\cite{controlnet} and 3D point cloud diffusion model~\cite{pointE}, it will be affected when ControlNet or Point-E fails with certain objects. This issue might be alleviated by developing more powerful foundation models.
Second, while achieving good controllability of 3D shapes, \Ours{} needs a single reference image for geometry guidance. This issue can be alleviated by cropping objects from real images using Segment Anything Model (SAM)~\cite{SAM}, or direct generating an image using text-to-image models, \eg, Stable Diffusion, ControlNet.

%################# Conclusions #################%
\section{Conclusions}
In this work, we propose \Ours{}, a novel and flexible text-to-3D generation framework. We inspire our framework by alleviating the view inconsistency problem and improving the controllability of 3D shapes for 3D content generation. 
To control the learned geometry, we innovatively propose to distill the geometry knowledge (sparse 3D points) from the 3D point cloud diffusion model (Point-E). To better utilize the sparse point cloud, we propose an efficient point cloud guidance loss to adaptively align the geometry between NeRF and sparse points.
Besides, to make the 3D content more realistic and view-consistent, we optimize the NeRF model conditioned on both text and the learned compact depth map, by performing score distillation to the 2D image diffusion model (ControlNet).
Both qualitative and quantitative comparisons demonstrate the superiority of \Ours{} in generating view-consistent and shape-controllable 3D contents.

%%
%% The next two lines define the bibliography style to be used, and
%% the bibliography file.
\bibliographystyle{ACM-Reference-Format}
\bibliography{sample-base}

%%
%% If your work has an appendix, this is the place to put it.
\appendix

\end{document}